# Survey of 3D Human Body Pose and Shape Estimation Methods for Contemporary Dance Applications

Darshan Venkatrayappa, Alain Tremeau, Damien Muselet and Philippe Colantoni
*Laboratoire Hubert Curien, Université Jean Monnet*
*18 Rue Professeur Benoît Lauras Bâtiment F, 42000 Saint-Étienne.*
*darsh.venkat@gmail.com,{alain.tremeau, damien.muselet, philippe.colantoni}@univ-st-etienne.fr*

Keywords: Human body shape estimation, Human body pose estimation, Contemporary Dance.

Abstract: 3D human body shape and pose estimation from RGB images is a challenging problem with potential applications in augmented/virtual reality, healthcare and fitness technology and virtual retail. Recent solutions have focused on three types of inputs: i) single images, ii) multi-view images and iii) videos. In this study, we surveyed and compared 3D body shape and pose estimation methods for contemporary dance and performing arts, with a special focus on human body pose and dressing, camera viewpoint, illumination conditions and background conditions. We demonstrated that multi-frame methods, such as PHALP, provide better results than single-frame method for pose estimation when dancers are performing contemporary dances.

## 1 INTRODUCTION

Capturing real-time human movement holds significance across various domains, including entertainment and gaming, fitness and sports, gesture and communication, healthcare, dance motion analysis and performing arts. While optical motion capture systems are the gold standard for precision, they rely on performers wearing sensor-equipped costumes or Motion Capture suits (MoCaps), posing challenges in theatrical contexts. Additionally, such systems can restrict dancers' freedom of movement. To address these issues, emerging approaches leverage machine learning and deep learning techniques. This survey aims to showcase the application of existing machine learning-based 3D human body shape and pose estimation models in the context of contemporary dance and performing arts. Humans are often a central element in images and videos. Understanding their posture, the social cues they communicate, and their interactions with the world is critical for holistic scene understanding. To understand human behaviour, however, we have to capture more than the major joints of the body, we need the full 3D surface of the body, hands and the face. 3D objects are often represented by vertices and triangles that encodes their 3D shape. The more detailed an object is, the more vertices are required. However, for human bodies the 3D mesh representation could be compressed down to a lower dimensional space whose axes are like their height, fatness, bust circumference, belly size, pose etc. This representation is often smaller and more meaningful. To represent human body in 3D, computer vision researchers have proposed optimization-based models like SMPL (Loper et al. 2015), SMPL-X (Pavlakos et al. 2019), FLAME (Li et al. 2017), STAR (Osman et al. 2020), and many more.

In the following sections, we will survey and compare 3D body shape and pose estimation methods for contemporary dance and performing arts, with a special focus on human body pose and dressing, camera viewpoint, illumination conditions and background conditions, tracking and occlusions.

## 2 OPTIMIZATION-BASED MODELS

### 2.1 SMPL-A Model

The SMPL (Skinned Multi-Person Linear) body model is a widely used computer graphics representation that aims to accurately model the 3D shape and pose of a human body. It provides a compact and parameterized representation of the body, enabling realistic animations and simulations. The model consists of a linear blend skinning approach, where a template mesh is deformed based on a set of pose and shape parameters. The pose parameters describe the joint rotations of the body, capturing movements such as bending or stretching of limbs. These parameters enable the animation of the model to simulate a wide range of natural human motions. The shape parameters, on the other hand, control the overall proportions and body fat distribution of the model, allowing for customization of individual body types. By combining the pose and shape parameters, the SMPL model allows for versatile control over the appearance and motion of virtual human characters, making it a valuable tool in various applications such as animation, virtual reality, and bio-mechanical simulations. The SMPL model offers benefits, including realistic body representations, efficiency for real-time applications, and wide availability. However, it has limitations such as limited facial and hand modelling and lacks representation of ethnic diversity. To enhance it, consider detailed facial and hand modelling, adaptability for diverse body shapes and ethnicities, customization, improved expressive poses, and optimization for real-time use, enabling integration into live performances and interactive experiences.

## 2.2 SMPL-X Model

The SMPL-X (Skinned Multi-Person Linear eXpression) body model builds upon its predecessor, the SMPL model, by addressing its limitations and introducing additional features. It offers a more comprehensive representation of the human body, encompassing facial expressions, hand poses, and improved joint accuracy. In the SMPL-X model, the body is depicted as a rigged mesh, combining a template mesh with associated joint positions and blend weights. This template mesh undergoes deformation based on pose and shape parameters, akin to the original SMPL model. However, the SMPL-X model extends these parameters to include facial expressions and hand poses, allowing for realistic modelling and animation of expressive facial expressions and hand gestures, enhancing the versatility of virtual characters.

Furthermore, the SMPL-X model enhances joint accuracy by incorporating data from 3D scans of actual human bodies. This integration aligns the model's joint positions more closely with anatomical landmarks, resulting in highly realistic and precise representations of the human body, as demonstrated in (Pavlakos et al. 2019). This increased precision proves especially valuable in applications like motion capture, virtual try-on experiences, and biomechanical simulations, where exact body alignment plays a pivotal role. Overall, the SMPL-X body model elevates the capabilities of the original SMPL model, enabling more realistic and detailed simulations of human bodies across a wide range of computer graphics and animation applications.

While the SMPL-X model brings several advantages, such as introducing facial expressions, basic hand gestures, and improved expressiveness, it also presents some limitations like simplified hand modelling and limited pose variability. To improve the SMPL-X model, potential enhancements can be considered. These include expanding its hand modelling capabilities to cover a broader range of hand gestures and interactions, integrating more intricate hand joint definitions, and resolving issues related to capturing extreme or unstable poses. These improvements aim to ensure that the model remains stable and realistic across a wide spectrum of poses.

## 2.3 MANO Model

MANO (Model for Articulated Hand Tracking) (Romero et al. 2017) is a computer graphics model designed to faithfully depict and simulate the 3D shape and movements of human hands. It offers a detailed and realistic portrayal of hand articulation, making it suitable for a range of applications, including hand tracking, animation, and interactions in virtual reality. The core of the MANO model comprises a parametric mesh that captures the intricate geometry of the hand, alongside a set of pose and shape parameters that govern how the hand moves and appears.

Within the MANO model, the pose parameters meticulously describe joint rotations in the hand, enabling precise control over the flexion, extension, and rotation of individual fingers and the wrist. This level of control facilitates the creation of realistic hand animations and interactions. Meanwhile, the shape parameters in MANO account for variations in hand geometry, encompassing aspects like finger lengths, palm width, and overall hand size. By manipulating these parameters, the model can faithfully represent diverse hand shapes and sizes. Additionally, MANO provides a comprehensive representation of the hand's surface, encompassing skin deformation and wrinkles, further enhancing the authenticity of virtual hand simulations. MANO finds practical utility across multiple domains, including virtual reality, and human-computer interaction, empowering realistic hand tracking, virtual hand animations, and interactive experiences involving hand gestures and manipulations.

While MANO offers benefits in the form of detailed hand gesture modelling and realistic hand deformations, it does have limitations, primarily related to its exclusive focus on hand modelling and the potential for computational challenges. To enhance the MANO model, there is room for improvement by prioritizing its ability to accurately represent intricate hand gestures, including finger articulations and precise movements. Additionally, there is an opportunity to explore methods for reducing the computational requirements of MANO while preserving its accuracy, making it more accessible and suitable for real-time applications.

## 2.4 FLAME Model

The FLAME (Fast Linear Albedo and Shape Model) model (Li et al. 2017) is an advanced computer graphics model engineered to meticulously capture the detailed shape, pose, and intricacies of human heads. This model excels in providing a highly realistic portrayal of facial geometry, encompassing fine-grained skin details, wrinkles, and expressions. It achieves this by skilfully integrating 3D face scans and a statistical learning framework. At its core, the FLAME model is comprised of two principal components: the shape model and the texture model. The shape model adeptly records the three-dimensional facial geometry, including the spatial arrangement of facial landmarks, cranial structure, and skin deformations. By manipulating the shape parameters, it affords precise control over facial expressions, shape variations, and head orientations. In contrast, the texture model captures facial appearance through the encoding of albedo and additional surface intricacies like wrinkles and pores, enabling realistic rendering and texturing of the face.

The FLAME model possesses key advantages, notably its computational efficiency, making it well-suited for real-time applications such as virtual reality, augmented reality, and video games. Its seamless integration into existing graphics pipelines empowers the creation of realistic and expressive virtual characters. This versatile model finds applications in diverse domains, including animation, virtual try-on, facial recognition, and biometrics, offering a powerful tool for generating and manipulating lifelike human faces in computer graphics and computer vision applications. However, it does have limitations, primarily focusing on facial and lip modelling, introducing complexity and potential computational challenges. Enhancements should prioritize optimizing real-time performance, especially when integrated with other components for full-body animations, and elevating detail and realism, particularly in soft tissues and articulations, to enhance overall animation quality.

## 2.5 STAR Model

The STAR (Sparse Temporal Articulated Regression) model (Osman et al. 2020) is a computer vision model designed for 3D human pose estimation from single or multiple video frames. It aims to accurately estimate the 3D positions of the body joints over time, enabling applications such as motion capture, action recognition, and human-computer interaction. The STAR model employs a sparse representation framework that utilizes both spatial and temporal information to robustly estimate the 3D pose of the human body. At its core, the STAR model combines two main components: a sparse coding module and a temporal fusion module. The sparse coding module extracts spatial features from individual video frames and encodes them into a compact representation. This module effectively captures the spatial relationships between body joints and local image features, enabling robust estimation of the 3D pose. The temporal fusion module utilizes the temporal coherence of human motion by incorporating the spatiotemporal information from neighbouring frames. This module ensures the consistency of the estimated poses over time and enhances the accuracy of the pose estimation.

The STAR model leverages the advantages of both sparse coding and temporal modelling, providing a powerful framework for 3D human pose estimation. It has demonstrated state-of-the-art performance in various benchmarks and has practical applications in domains such as sports analysis, human-computer interaction, and virtual reality. By accurately estimating the 3D pose from video data, the STAR model contributes to a wide range of applications that require understanding and analysis of human motion.

## 2.6 Application of optimization-based 3D models in dance/theater

The SMPL model and its derivatives offer potent capabilities within the realm of dance and theater. While initially designed for computer graphics and computer vision applications, specifically 3D human pose and shape estimation from 2D imagery and videos, in this context, they serve to fit 3D models to archived videos. Additionally, they excel in creating lifelike 3D character animations, which prove invaluable for bringing virtual characters to life on stage or in digital productions within the dance and theater domain. Augmented with motion capture data, these models accurately depict the movements of actors and dancers, aiding in choreography development, rehearsals, and precise movement analysis. Moreover, they support costume designers by estimating performers' body shapes and dimensions, enabling custom-fitted outfits for enhanced aesthetics. In theater and dance auditions, these models provide pre-visualization tools to assist directors in casting decisions.

In dance performances, MANO finds purpose in recognizing and interpreting intricate hand gestures and movements, offering insights into dancers' expressive choreography. It can seamlessly integrate with lighting and effects systems to facilitate gesture-controlled adjustments of stage elements, injecting dynamism into performances. For injury rehabili-tation or therapeutic dance and theater programs, MANO tracks and analyses hand movements. Furthermore, it facilitates hand-based interactions between performers and digital elements, crafting immersive and interactive theater experiences.

FLAME's renowned highly detailed facial model lends itself to creating realistic facial animations. In theater, it elevates storytelling and character portrayal by animating digital characters or avatars. Its facial modelling capabilities extend to aiding makeup artists and costume designers in crafting custom makeup and prosthetics that precisely match actors' facial features, ensuring seamless character integration. FLAME's integration with facial tracking technologies captures and analyses actors' facial expressions during live performances, enabling real-time adjustments of digital characters or lighting effects.

In the realm of virtual theater and dance productions, SMPL-X, along with the FLAME model, creates virtual avatars or characters mirroring the movements and expressions of real actors or dancers. This is particularly valuable for immersive experiences and remote performances. Combining SMPL-X, FLAME, and MANO serves as an educational tool in dance and theater instruction, enhancing students' understanding of body movement and anatomy, ultimately refining their performance skills. In modern theater and dance productions that frequently employ digital effects and projections, these 3D models contribute to crafting realistic digital effects, offering precise human body representations for interaction with virtual elements. These models seamlessly integrate into interactive and augmented reality performances, where performers' movements are tracked to control digital elements in real-time.

The examples listed above are typical applications addressed in the EU project PREMIERE - Performing arts in a new area (Premiere 2023).

## 3 DEEP LEARNING-BASED MODELS

The optimization-based 3D human body models mentioned in the previous section offer meticulous control and robust mathematical foundations, but they come with a drawback of being computationally demanding and less suitable for real-time usage. In contrast, deep learning-based models deliver accuracy driven by data, versatility, and real-time capabilities, making them particularly well-suited for applications like animation, virtual reality, and interactive experiences. The decision between these two model types hinges on the specific demands of the application, the available computational resources, and the trade-offs between precision and efficiency. Within the context of performing arts, prioritizing accurate modelling of human body motion patterns takes precedence over achieving realistic 3D renderings of human body shapes or corresponding avatars. In the following section, we outline some of the deep learning-based methods for 3D body and shape estimation.

### 3.1 HMR Model

Human Mesh Recovery (HMR) is a widely-used top-down, end-to-end framework for reconstructing a full 3D mesh of a human body from a single RGB image (Kanazawa et al. 2018). A square cropped image is resized to 224x224 and passed through a convolution encoder. In contrast to most of the methods that compute 2D or 3D joint locations, HMR produce a richer and more useful mesh representation that is parameterized by shape and 3D joint angles. The main objective of HMR is to minimize the re-projection loss of key-points, which allows the model to be trained using in-the-wild images that only have ground truth 2D annotations.

However, the re-projection loss alone is highly under-constrained. HMR address this problem by introducing an adversary trained to tell whether human body shape and pose parameters are real or not using a large database of 3D human meshes. The idea is that, given an image, the network has to infer the 3D mesh parameters and the camera such that the 3D key-points match the annotated 2D key-points after projection. To deal with ambiguities, these parameters are sent to a discriminator network, whose task is to determine if the 3D parameters correspond to bodies of real humans or not. Hence the network is encouraged to output parameters on the human manifold and the discriminator acts as weak supervision. The network implicitly learns the angle limits for each joint and is discouraged from making people with unusual body shapes.

## 3.2 VIBE Model

Human motion is fundamental to understanding behaviour. Despite progress on single-image 3D pose and shape estimation, existing video-based state-of-the-art methods fail to produce accurate and natural motion sequences due to a lack of ground-truth 3D motion data for training. To address this problem authors of (Kocabas et al. 2020) propose VIBE (Video Inference for Human Body Pose and Shape Estimation). VIBE uses CNNs, RNNs (GRU) and GANs along with a self-attention layer to achieve its state-of-the-art results. VIBE uses CNNs to extract image features. The output from the CNN is fed as input to the recurrent neural network, which processes the sequential nature of human motion. Then, a temporal encoder and regressor are used to predict the body parameters for the whole input sequence. This whole part is referred to as the Generator(G) model. Now with the help of the AMASS dataset 3D, realistic human motion is achieved for adversarial training and build a motion discriminator (Dm). The motion discriminator takes in both predicted pose sequences along with pose sequences sampled from AMASS. The discriminator tries to differentiate between the fake and real motions by providing a real/fake probability for each input sequence which helps in producing realistic motion. The output of this method is a standard SMPL body model format consisting sequence of pose and shape parameters.

## 3.3 SPIN Model

In the computer vision literature, the model-based human pose estimation is currently approached through optimization-based methods and regression-based methods. Optimization-based methods fit a parametric body model to 2D observations in an iterative manner, leading to accurate image model alignments, but are often slow and sensitive to the initialization. On the other hand, regression-based methods, that use a deep network to directly estimate the model parameters from pixels, tend to provide reasonable, but not pixel accurate, results while requiring huge amounts of supervision. The authors of (Kolotouros et al. 2019) propose SPIN (SMPL oPtimization IN the loop), a self-improving approach for training a neural network for 3D human pose and shape estimation, through the tight collaboration of a regression- and an optimization-based method.

Instead of using the ground truth 2D keypoints to apply a weak re-projection loss, the authors propose to use regressed estimate to initialize an iterative optimization routine that fits the model to 2D keypoints. This procedure is done within the training loop. The optimized model parameters are used to explicitly supervise the output of the network and supply it with privileged model-based supervision, that is beneficial compared to the weaker and typically ambiguous 2D reprojection losses. This collaboration leads to a self-improving loop, since better fits help the network train better, while better initial estimates from the network helps the optimization routine converge to better fits.

## 3.4 PARE Model

State of the art 3D human pose and shape estimation methods remain sensitive to partial occlusion and can produce dramatically wrong predictions although much of the body is observable. Authors of (Kocabas et al. 2021) address this by introducing a soft attention mechanism, called PARE. PARE (Part Attention Regressor for 3D Human Body Estimation) learns to predict body-part-guided attention masks. Most of the state-of-the-art methods rely on global feature representations, making them sensitive to even small occlusions. In contrast, PARE's part-guided attention mechanism overcomes these issues by exploiting information about the visibility of individual body parts while leveraging information from neighbouring body-parts to predict occluded parts.

PARE has two tasks: the primary one is learning to regress 3D body parameters in an end-to-end fashion, and the auxiliary task is learning attention weights per body part. Each task has its own pixel-aligned feature extraction branch $P$ and $F$. which are fused by part attention leading to the final feature $F'$ for camera and SMPL body regression. Here the key insight is that, to be robust to occlusions, the network should leverage pixel-aligned image features of visible parts to reason about occluded parts.

### 3.5 EXPOSE Model

Accurate and fast prediction of the 3D body, face, and hands together from an RGB image is an important aspect in understanding how people look, interact, or perform tasks. Current methods focus only on parts of the body. A few recent approaches reconstruct full expressive 3D humans from images using 3D body models that include the face and hands. Most of the methods are optimization-based and thus slow, prone to local optima, and require 2D keypoints as input. The authors of (Choutas et al. 2020) address these limitations by introducing ExPose (EXpressive POse and Shape rEgression), which directly regresses the body, face, and hands, in SMPL-X format, from an RGB image. This is a hard problem due to the high dimensionality of the body and the lack of expressive training data. Additionally, hands and faces are much smaller than the body, occupying very few image pixels. Initially, the authors account for the lack of training data by curating a data-set of SMPL-X fits on in-the-wild images. Secondly, as body estimation localizes the face and hands reasonably well. They propose body-driven attention for face and hand regions in the original image to extract higher-resolution crops that are fed to dedicated refinement modules. Finally, these modules exploit part-specific knowledge from existing face and hand-only data-sets. ExPose estimates expressive 3D humans more accurately than existing optimization methods at a small fraction of the computational cost.

An image of the body is extracted using a bounding box from the full resolution image and fed to a neural network $g(.)$, that predicts body pose, hand-pose, facial pose, shape, expression, camera scale and translation. Face and hand images are extracted from the original resolution image using bi-linear interpolation. These are fed to part specific sub-networks $f(.)$ and $h(.)$, respectively to produce the final estimates for the face and hand parameters. During training the part specific networks can also receive hand and face only data for extra supervision.

### 3.6 PHALP Model

The PHALP method (Rajasegaran et al. 2022) presents an approach for tracking people in monocular videos by predicting their future 3D representations. The first step of this method is to lift people to 3D from a single frame in a robust way, which includes information about the 3D pose of the person, their location in the 3D space, and the 3D appearance. As they track a person, they collect 3D observations over time in a tracklet representation. Given the 3D nature of the observations, they build temporal models for each one of the previous attributes and use these models to predict the future state of the tracklet, including 3D location, 3D appearance, and 3D pose. For a future frame, they compute the similarity between the predicted state of a tracklet and the single frame observations in a probabilistic manner. Association is solved with simple Hungarian matching, and the matches are used to update the respective tracklets.

One of the main limitations of the PHALP method is that it relies on a single camera to capture the video, which can lead to issues such as occlusion, where the person being tracked is partially or completely hidden from view, or motion blur, where the person's appearance is distorted due to rapid movement. Additionally, the method may not work well in low-light conditions or when the person is wearing clothing that is similar in color or texture to the background. These are some of the factors that can affect the performance of the method. It is also worth noting that the method requires a significant amount of computational resources, which can make it difficult to implement in real-time applications. Furthermore, the method may not be suitable for tracking people in crowded environments, where there are multiple people moving in close proximity to each other.

## 4 EXPERIMENTAL RESULTS

In our experiments, we associated a Grade (G) and a Category (C) to each processed image, as there are various types of estimation errors. Note that a processed image may suffer from several types of estimation errors. We considered five estimation error grades: 5 - severe distortions; 4 - strong distortions; 3 - minor distortions; 2 - realistic 3D model; 1- accurate 3D model. In the following, the "best grade" among methods will be indicated in bold. When we have more than one image with "best grade", only the best image is indicated in bold. When the "best grade" is upper than 3 we consider that no method performs well. We considered five categories of estimation errors: 0. no noticeable error; 1. miss alignment (of the full body or of some body parts); 2. miss estimation of the body shape (size, height, width, orientation, etc.); 3. missing 3D model or ghosted 3D model; 4. incoherency of limbs positions.

Figures 1 and 2 illustrate the limits of HMR, VIBE, SPIN and PARE 3D human body pose estimation methods, especially on limbs. Sometimes the limbs are not aligned with the 3D body model. The miss-alignments are worse for Figure 1 than for Figure 2 when the range of movements is higher. Miss-alignments are worse for the hands (when the frame rate of the video sequence is lower than the movement speed[a], in such case we have motion blur) and for the feet (when the angle between the feet and the leg is too strong). The PARE method performs better than HMR, VIBE and SPIN when applied to the 14$^{th}$ video sequence of the Talawa video (Tabanka 2018), see Figure 1, the 3D models are better aligned with the human body of dancers (this can be observed on the heads). The 3D models computed using the PARE method are comparable with the ones computed using VIBE, and better than the ones computed using HMR or SPIN, when applied to the 15$^{th}$ video sequence of the Talawa video, see Figure 2, the 3D models are better aligned with the human body of dancers (this can be observed on the heads).

For the following experiments, we used another set of videos from the "PREMIERE Dance Motion Dataset"[b], specifically designed to evaluate the accuracy and robustness of 3D pose estimation methods (Premiere 2023). In the following, to demonstrate that single-frame pose estimation methods perform worse than multi-frame -based methods, such as PHALP (Rajasegaran et al. 2022), we will show results obtained using the PHALP method. This method combines 3D pose estimation from single-frame with location into the 3D space and appearance information for each person over multiple frames. It also aggregates single frame representa-tions over time, predicts future representations, and associates tracks with detections using predicted representations in a probabilistic framework.

Figures 3 and 4 illustrate the limits of PHALP, PARE, VIBE and SPIN 3D human body pose estimation methods, especially for PARE, VIBE and SPIN on limbs. The miss-alignments are worse for Figure 4 than for Figure 3, when the range of movements is higher. The last rows of Figure 4 demonstrate than the miss-alignments increase when the angle between the left and the right legs of the dancers increase, except for the PHALP method which performs well in all cases. Results shown in Fig. 4 show that PHALP performs better than all other methods. Results shown in Fig. 3 show that PHALP performs better than all other methods, next the best results are obtained with VIBE (less miss-alignments on limbs than with PARE and SPIN). The worse results are those obtained with SPIN.

Figures 5, 6 and 7 illustrate the limits of body pose estimation methods when we have strong occlusions between dancers. In all video sequences shown in Figures 5 to 7, PHALP is able to well estimate the pose of the front dancer and for few image frames faces issues to detect the second dancer in the back of the front dancer, especially when the legs of the second dancer are not visible during few consecutive frames (see Figure 5). In Figure 6, the pose estimation is good for the two dancers, even if we have strong occlusion between dancers, as the legs of the second dancer in the back of the front dancer are visible for most of the frames. There is less occlusion in the video sequence shown in Figure 7, so the pose estimation is good for the two dancers. PARE, VIBE and SPIN methods fail to detect properly each dancer in all video sequences shown in Figures 5 to 7. The pose estimation is worse for the video sequence shown in Figure 3, as there is a confusion between the upper part of the second dancer in the back of the front dancer and the legs of the front dancer, consequently the 3D model estimated is incorrect. Same kind of confusion can be observed in Figure 6. There is no confusion in Figure 7, results of PARE, VIBE and SPIN methods are quite comparable.

Several metrics can be used to evaluate the accuracy of pose estimation methods, such as the Percentage of Detected Joints (PDJ) for the whole body, the Object Keypoint Similarity (OKS) for each body part, the Percentage of Correct Parts (or Correctly Estimated Body Parts) – PCP for each body part of a set of images, Percentage of Correct Key-points - PCK, the Mean Per Joint Position Error - MPJPE, and the Multi-Instance Mean Squared Error (Khirodkar et al. 2021). These metrics require to know the true position of keypoints (i.e. to have access to geometrically calibrated images). In our experiments, we demonstrated that occlusions have a strong impact on the accuracy of pose estimation, even if tracking methods could be used to smooth temporal occlusions through multi-frames (as in Figures 5, 6 and 7). Results reported in this survey demonstrate that torso viewpoint, part length (foreshortening) and activity have also an impact on the accuracy (as in Figure 3). Results reported in this survey demonstrate also that, for dance motion analysis, the accuracy of pose estimation is less important than the analysis of dance motion patterns (as in Figure 3).

---

[a] The frame rate for the Talawa video (Tabanka 2018) is of 25 FPS. The low resolution of frames has also an impact on the accuracy of body parts pose estimation (each frame is of 1280x720 pixels).

[b] The frame rate for the PREMIERE dance motion dataset (Premiere 2023) is of 60 FPS. The resolution of image frames is of 1280x720 pixels.

Lastly, the limits of UpToDate pose estimation methods have been demonstrated by detection errors that occur when these later are applied to complex video contents, such as those put forward by the "PREMIERE dance motion dataset".

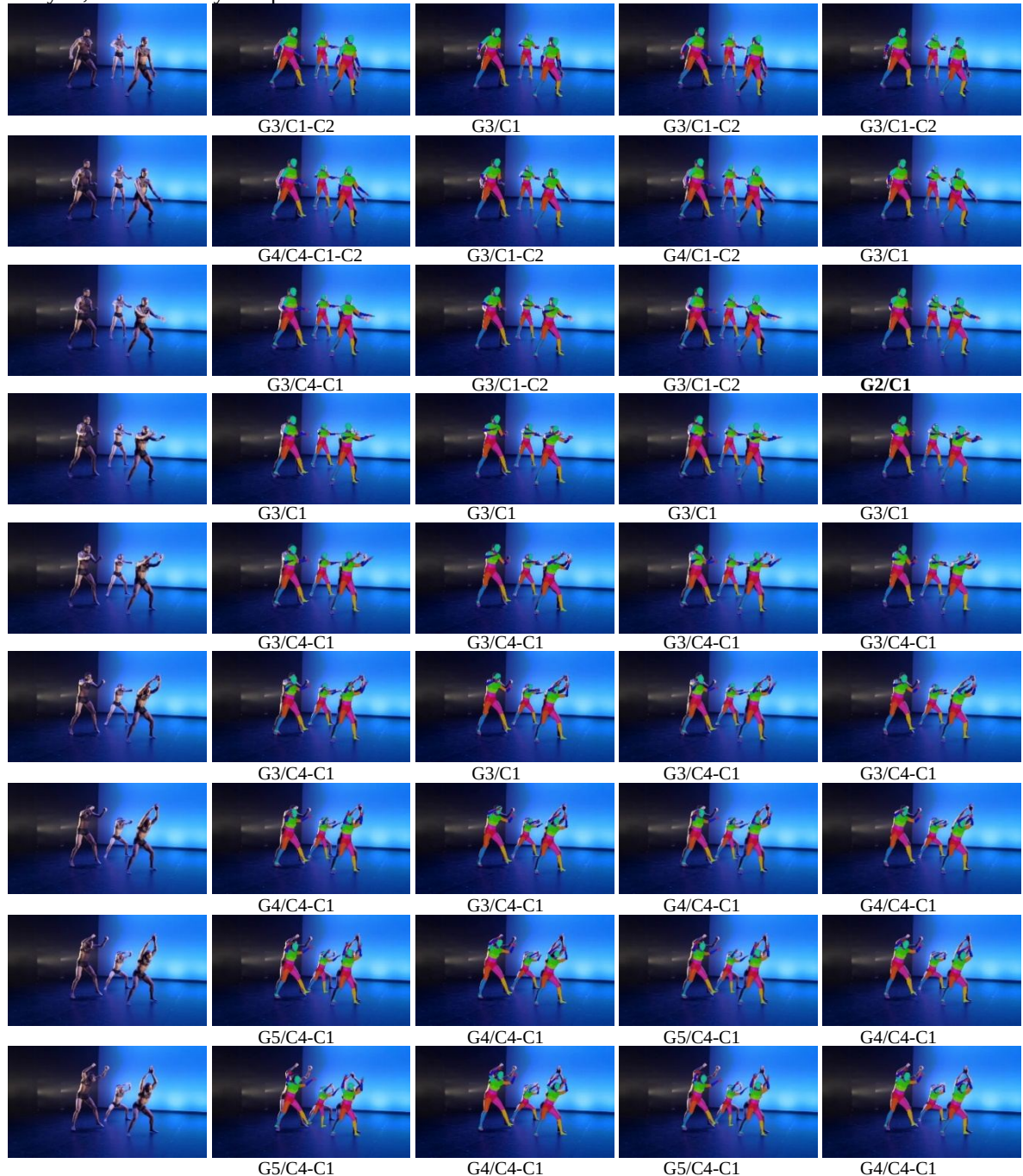

| | | | |
|---|---|---|---|
| G3/C1-C2 | G3/C1 | G3/C1-C2 | G3/C1-C2 |
| G4/C4-C1-C2 | G3/C1-C2 | G4/C1-C2 | G3/C1 |
| G3/C4-C1 | G3/C1-C2 | G3/C1-C2 | **G2/C1** |
| G3/C1 | G3/C1 | G3/C1 | G3/C1 |
| G3/C4-C1 | G3/C4-C1 | G3/C4-C1 | G3/C4-C1 |
| G3/C4-C1 | G3/C1 | G3/C4-C1 | G3/C4-C1 |
| G4/C4-C1 | G3/C4-C1 | G4/C4-C1 | G4/C4-C1 |
| G5/C4-C1 | G4/C4-C1 | G5/C4-C1 | G4/C4-C1 |
| G5/C4-C1 | G4/C4-C1 | G5/C4-C1 | G4/C4-C1 |

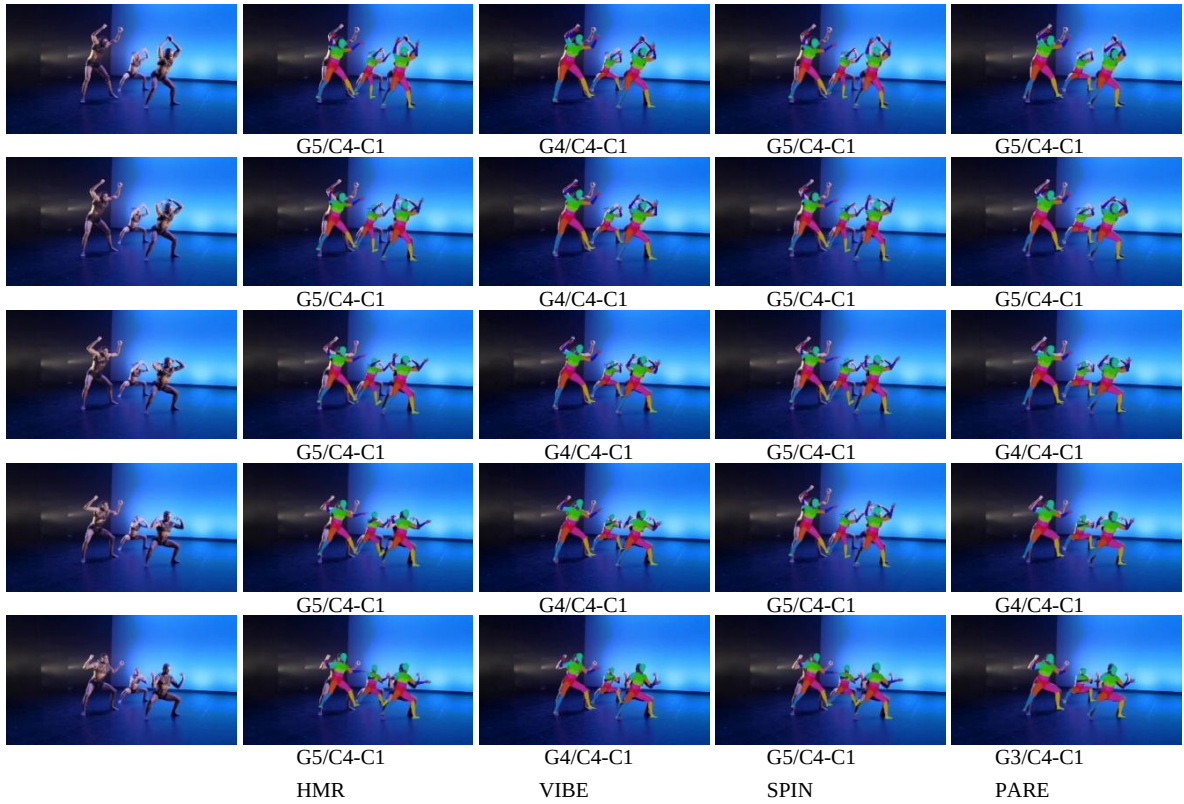

Figure 1: Frames in the 1st column represent the original frames of the 14th video sequence of the Talawa video. Comparison of 3D Human body pose estimation results, from the 2$^{nd}$ to the 5$^{th}$ column results from HMR, VIBE, SPIN an PARE method, respectively.

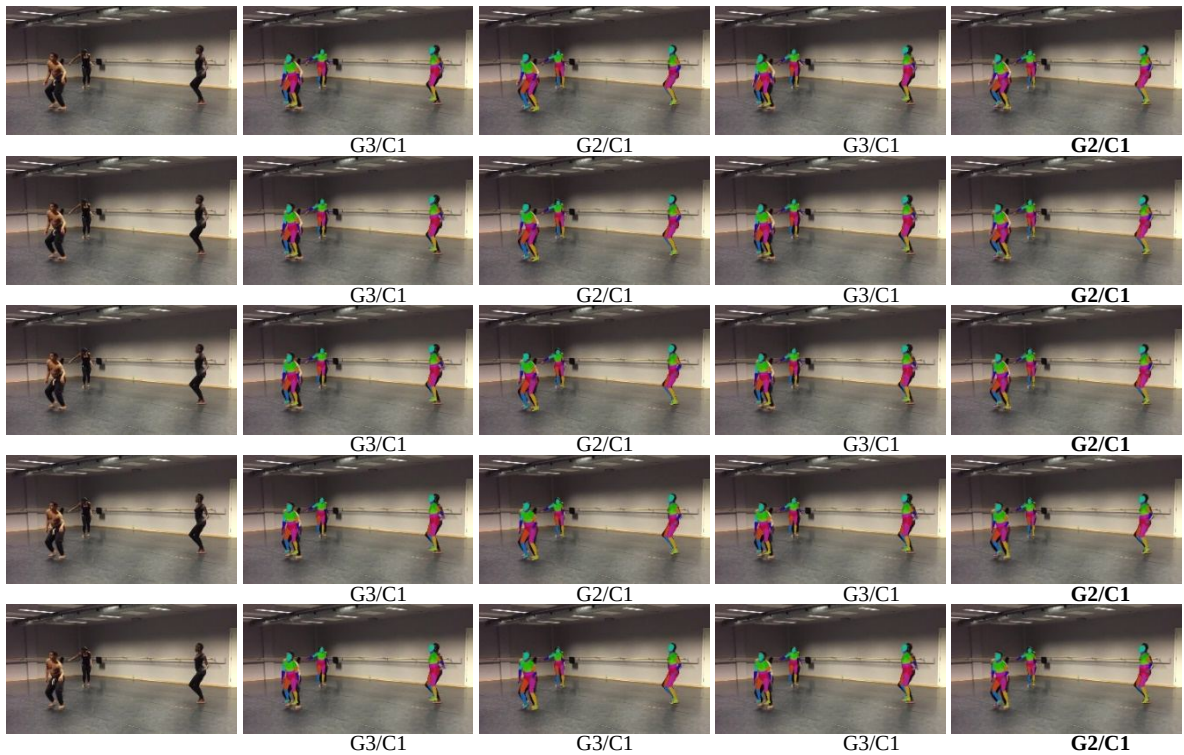

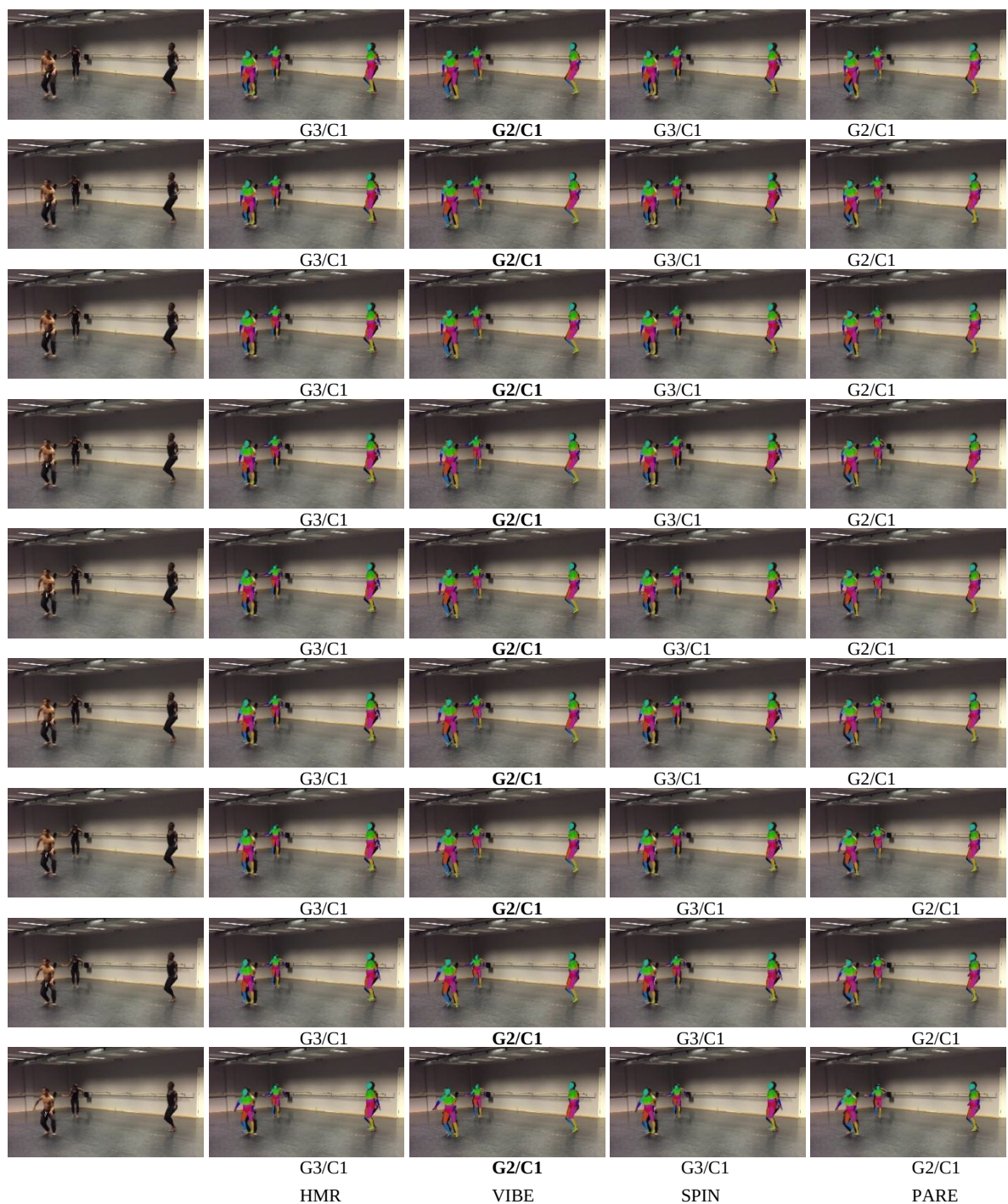

Figure 2: Frames in the 1st column represent the original frames of the 15th video sequence of the Talawa video. Comparison of 3D human body pose estimation results, from the 2$^{nd}$ to the 5$^{th}$ column results from HMR, VIBE, SPIN an PARE method, respectively.

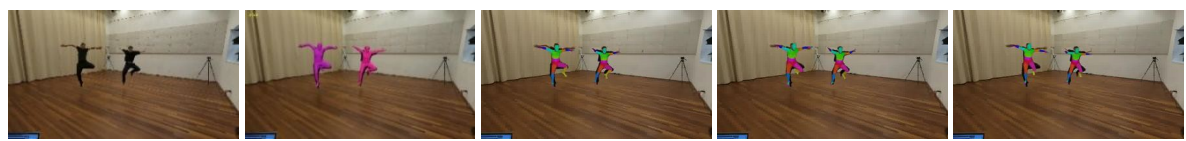

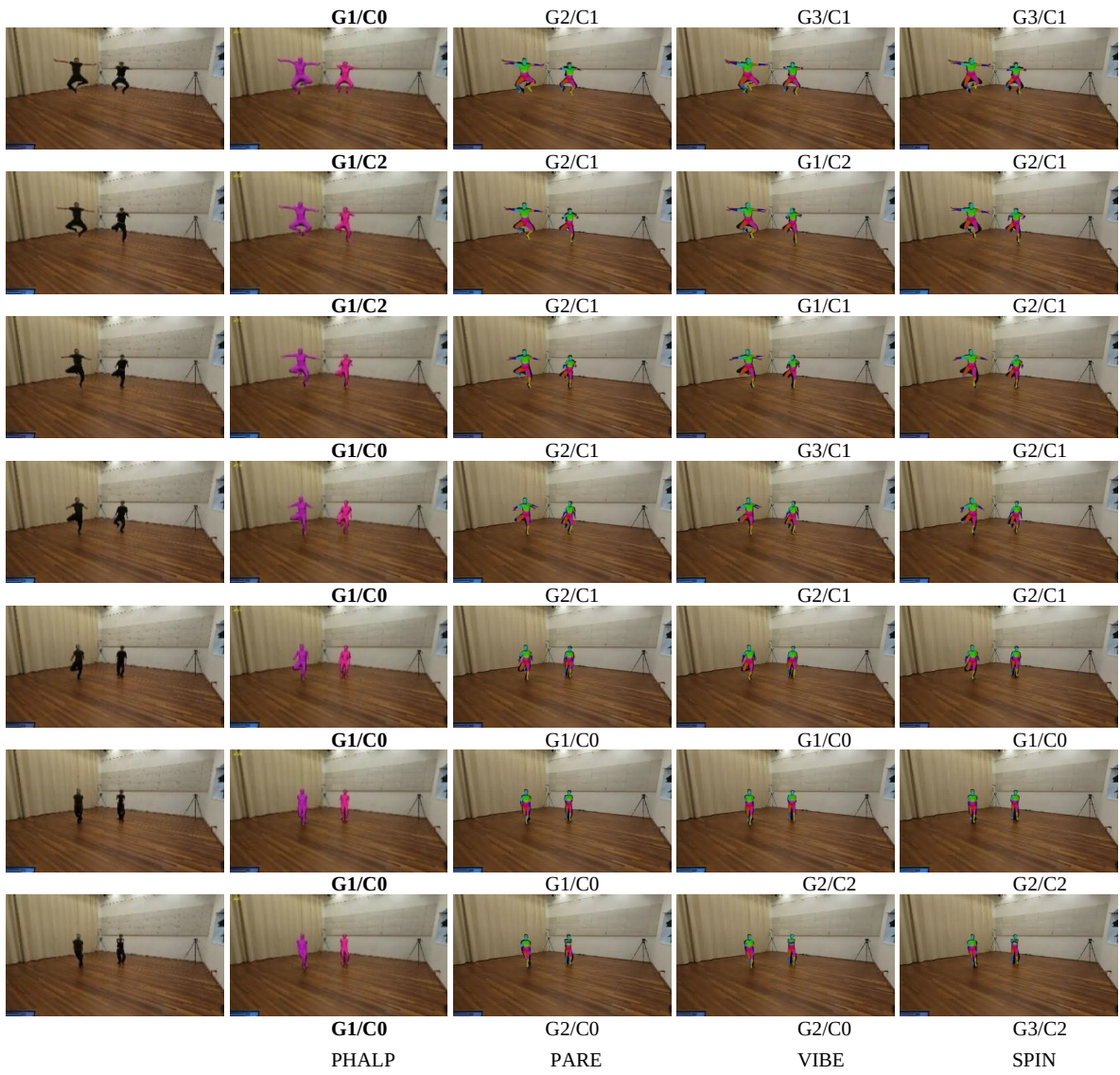

Figure 3: Frames in the 1st column represent the original frames of one video sequence of the PREMIERE Dance Motion Dataset with two dancers and no occlusion. Comparison of 3D Human body pose estimation results, from the 2$^{nd}$ to the 5$^{th}$ column results from PHALP, PARE, VIBE and SPIN method, respectively.

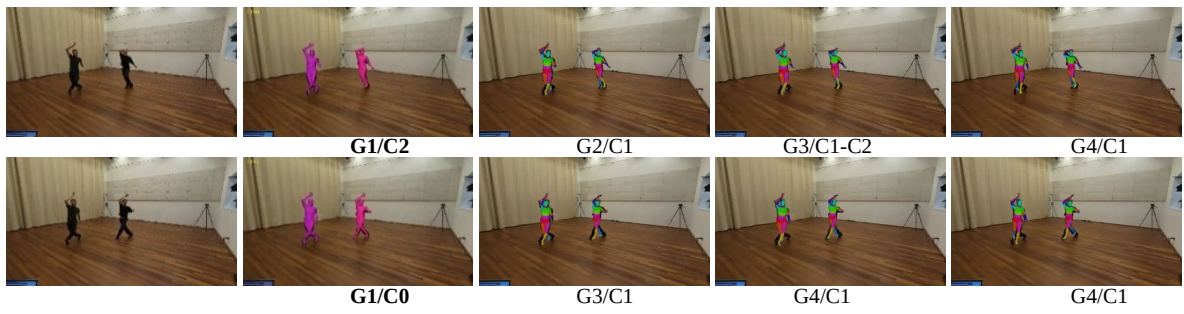

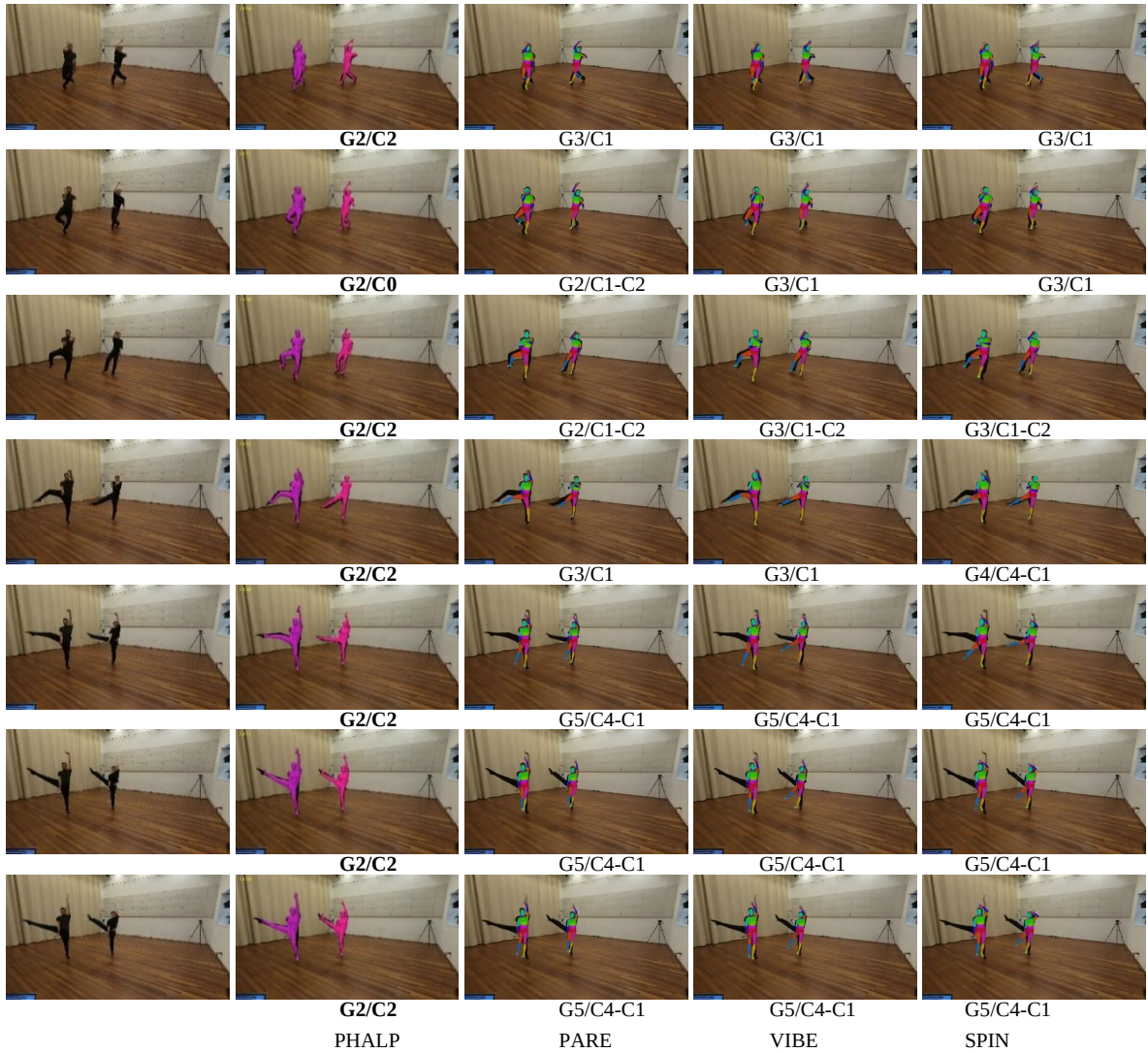

Figure 4: Frames in the 1st column represent the original frames of another video sequence of the PREMIERE Dance Motion Dataset with also two dancers and no occlusion. Comparison of 3D Human body pose estimation results, from the 2nd to the 5th column results from PHALP, PARE, VIBE and SPIN method, respectively.

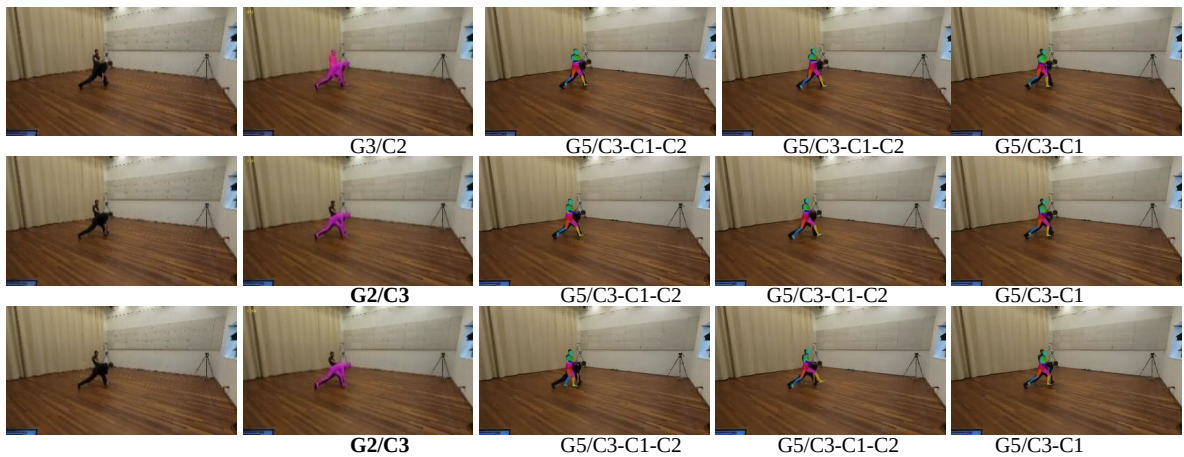

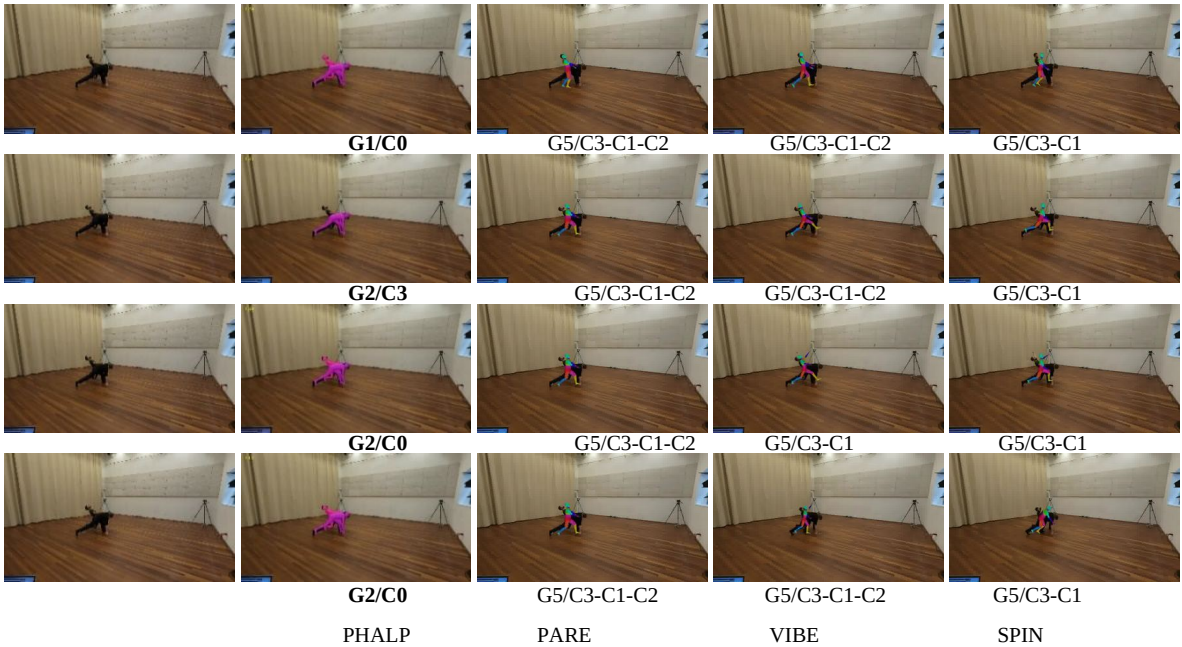

| | PHALP | PARE | VIBE | SPIN |

Figure 5: Frames in the 1st column represent the original frames of another video sequence of the PREMIERE Dance Motion Dataset with two dancers and strong occlusions. Comparison of 3D Human body pose estimation results, from the 2$^{nd}$ to the 5$^{th}$ column results from PHALP, PARE, VIBE and SPIN method, respectively.

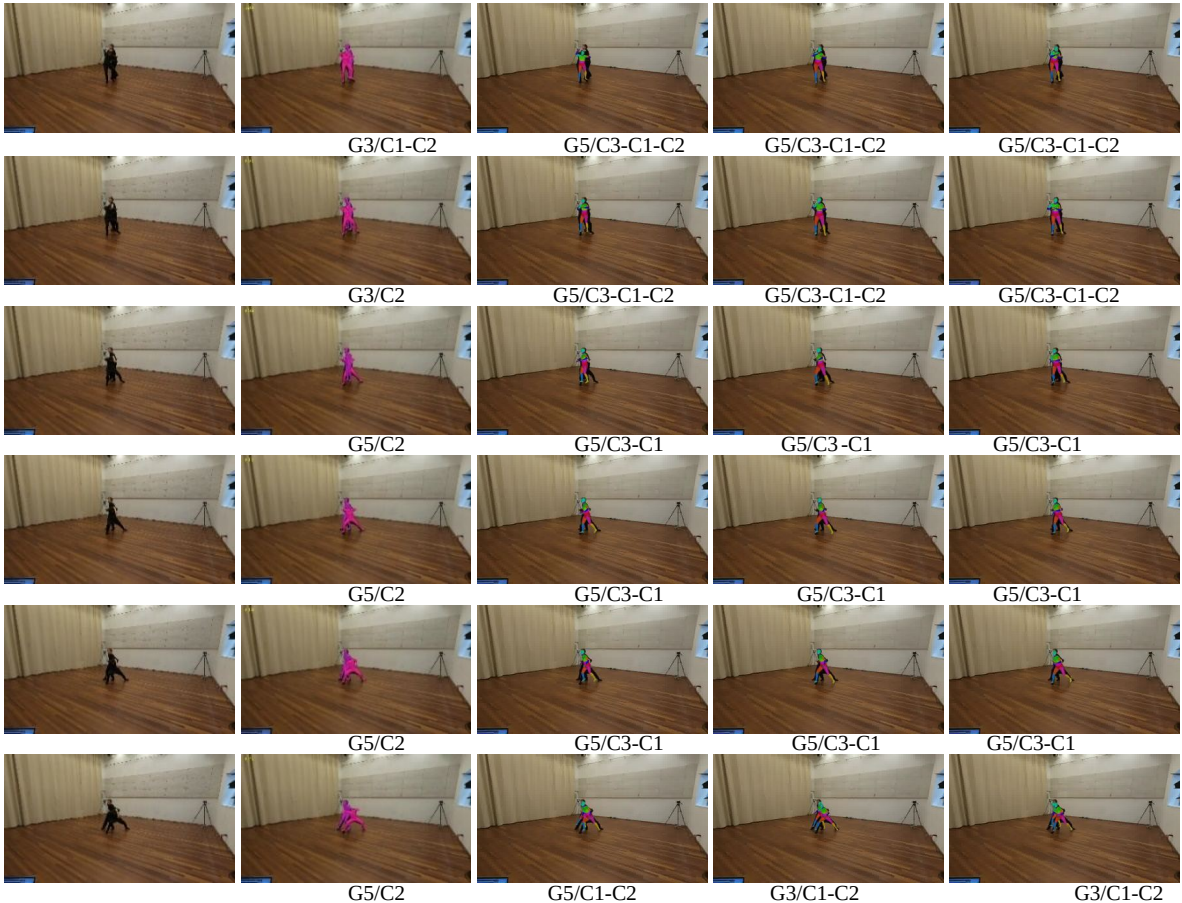

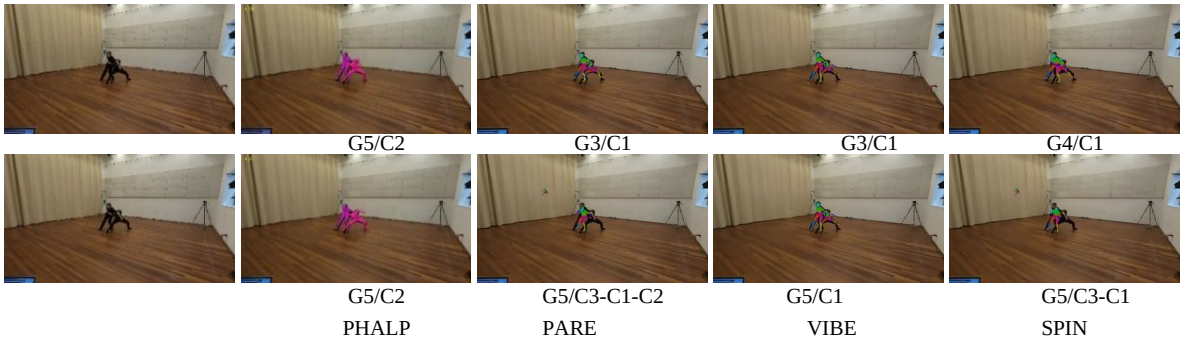

|  | G5/C2 | G3/C1 | G3/C1 | G4/C1 |
|---|---|---|---|---|
|  | G5/C2 | G5/C3-C1-C2 | G5/C1 | G5/C3-C1 |
|  | PHALP | PARE | VIBE | SPIN |

Figure 6: Frames in the 1st column represent the original frames of another video sequence of the PREMIERE Dance Motion Dataset with two dancers and strong occlusions. Comparison of 3D Human body pose estimation results, from the 2nd to the 5th column results from PHALP, PARE, VIBE and SPIN method, respectively.

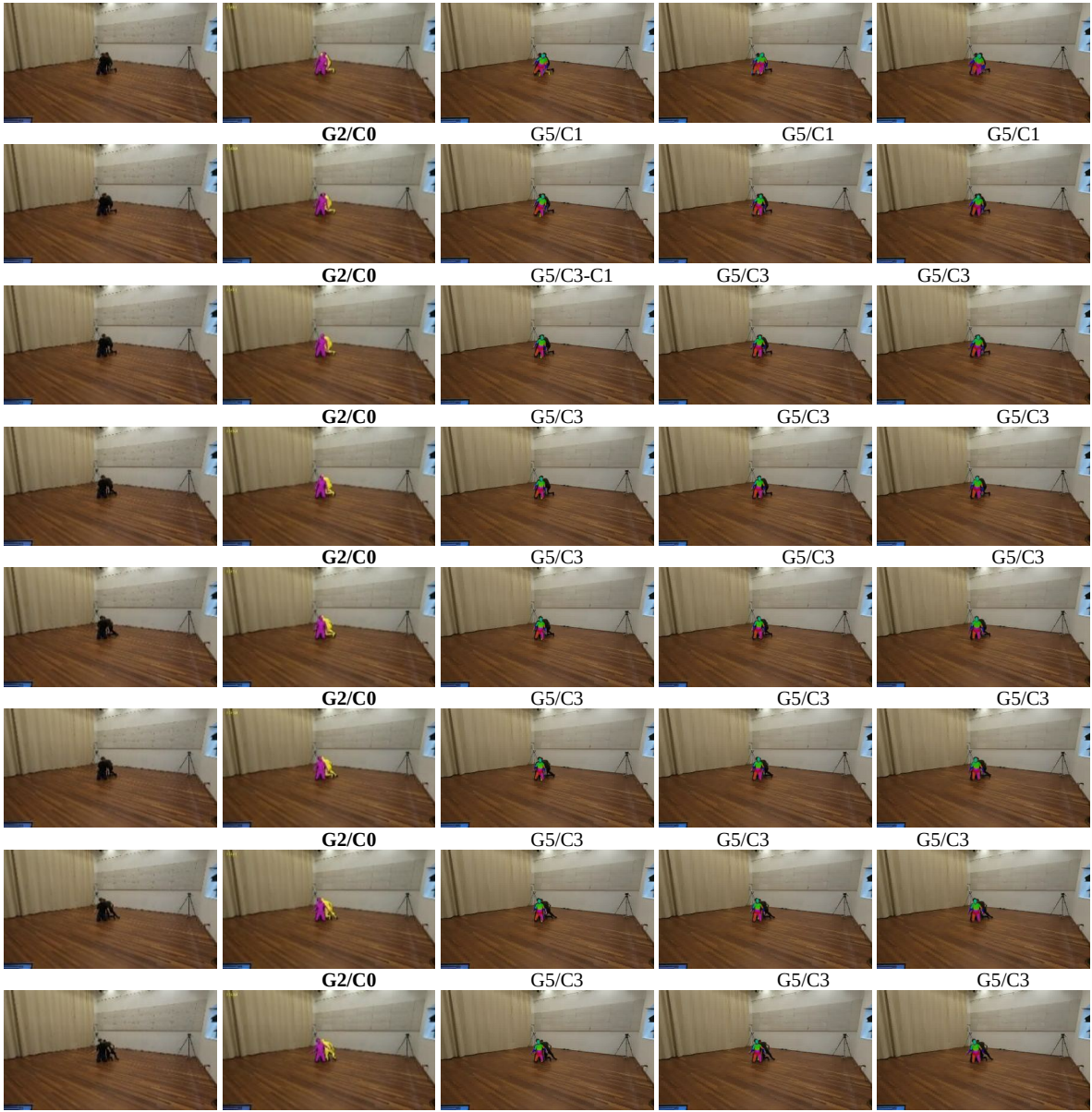

| | | | |
|---|---|---|---|
| **G2/C0** | G5/C3 | G5/C3 | G5/C3 |
| PHALP | PARE | VIBE | SPIN |

Figure 7: Frames in the 1st column represent the original frames of another video sequence of the PREMIERE Dance Motion Dataset with two dancers and strong occlusions. Comparison of 3D Human body pose estimation results, from the 2nd to the 5th column results from PHALP, PARE, VIBE and SPIN method, respectively.

# 4 CONCLUSIONS

In this survey, we discussed about the applications of 3d human body shape and pose models and their application to contemporary dance and performing arts. From the qualitative results it can be seen that the methods perform well when there is no occlusion or when the movement/motion is not too fast. Additionally, one of the major limitations of these methods is the real time performance.

We demonstrated that the state-of-the-art human body pose estimation models are not sufficiently well designed for contemporary dance application. To face this issue, we suggest the following improvements. Firstly, these methods need to be retrained using an appropriate dance dataset (this is what we are currently doing). Secondly, the state-of-the-art 3D human body models need to be improved to better consider unconventional body shape position and motion patterns specific to the dance domain. Thirdly, 2D pose estimation models could be refined from data provided by other 2D views of the same scene (some body parts are better estimated from front view, meanwhile others are better estimated from perpendicular view). Lastly, some geometrical constraints could be added to the current models (e.g. feet should not be below the ground).

We assume that for dance motion analysis the accuracy of human body pose estimation is less important for dance pattern analysis or avatars animations than the realism of the motion patterns. We will investigate this in the next future.

# ACKNOWLEDGEMENTS

This work was supported by HORIZON-CL2-2021-HERITAGE-01-04 grant. Project: 101061303 — PREMIERE. The two dancers of the "PREMIERE dance motion dataset" are [Jeroen Janssen](#) and [Rodrigo Ribeiro](#), they performed at [AHK ID-Lab](#) at Amsterdam on 21 Sept. 2023.